\documentclass[letterpaper, 10 pt, conference]{ieeeconf}  

\IEEEoverridecommandlockouts                              

\usepackage{graphics}
\usepackage{epsfig}
\usepackage{color}
\usepackage{tabularx}
\usepackage{makecell}
\usepackage{caption}
\usepackage{adjustbox}
\usepackage{array}
\usepackage{times}
\usepackage{amsmath}
\usepackage{amssymb}
\usepackage{afterpage}
\usepackage{algorithmic}
\usepackage[utf8]{inputenc}
\usepackage[normalem]{ulem}
\usepackage{multicol}
\usepackage{graphicx}
\usepackage{textcomp}
\usepackage{subfigure}
\usepackage{capt-of}
\usepackage{epstopdf}
\usepackage{lipsum}

\usepackage{amsmath} 
\usepackage{amssymb}  
\usepackage{epstopdf}
\usepackage{lipsum}
\usepackage{lipsum,amsmath,multicol}
\usepackage{float}
\usepackage[linesnumbered,ruled,vlined]{algorithm2e}
\SetKwInput{KwInput}{Input}        
\SetKwInput{KwReturn1}{Return}    
\SetKwInput{KwOutput}{Output} 
\SetKwInput{KwInitialization}{Initialization} 
\usepackage{wrapfig}
\usepackage{sidecap}

\usepackage{amsthm}

\usepackage{comment}
\usepackage{array}
\usepackage{glossaries}
\usepackage{breqn}
\usepackage{mathtools, cuted}
\makeglossaries
\newcolumntype{L}[1]{>{\raggedright\let\newline\\\arraybackslash\hspace{0pt}}m{#1}}
\newcolumntype{C}[1]{>{\centering\let\newline\\\arraybackslash\hspace{0pt}}m{#1}}
\newcolumntype{R}[1]{>{\raggedleft\let\newline\\\arraybackslash\hspace{0pt}}m{#1}}
\usepackage{sidecap}
\usepackage{url}
\usepackage[T1]{fontenc}
\usepackage{tabularx}
\usepackage{multicol, blindtext}
\usepackage{amsmath}
\usepackage{multirow}

\usepackage[table,xcdraw]{xcolor}
\usepackage{epsfig}         
\usepackage{times}          
\usepackage{amsmath}        
\usepackage{amssymb}        
\usepackage{afterpage}
\usepackage{makecell}
\usepackage{tabularx}
\usepackage{algorithmic}
\usepackage{caption}
\usepackage{array}
\usepackage{adjustbox}
\usepackage[utf8]{inputenc}
\usepackage[normalem]{ulem}
\usepackage{multicol}
\usepackage{hyperref}
\usepackage{cleveref}
\usepackage{cancel}
\usepackage[style=ieee]{biblatex} 
\usepackage{dblfloatfix}
\def\equationautorefname~#1\null{(eq~#1)\null}
\def\subsectionautorefname~#1\null{(sec~#1)\null}

\title{\LARGE \bf
GPD: \uline{G}uided \uline{P}olynomial \uline{D}iffusion for Motion Planning

}

\author{Ajit Srikanth$^{*1}$, Parth Mahajan$^{*1}$, Kallol Saha$^{*1,2}$, Vishal Mandadi$^{*1}$, Pranjal Paul$^{1}$, Pawan Wadhwani$^{1}$ \\
        Brojeshwar Bhowmick$^{3}$, Arun Singh$^{4}$, and Madhava Krishna$^{1}$
\thanks{* Denotes equal contribution}
\thanks{$^{1}$ Robotic Research Center, IIIT Hyderabad, $^{2}$ Carnegie Mellon University, $^{3}$ TCS Research, $^{4}$ University of Tartu}
\thanks{Project website: \href{https://guided-polynomial-diffusion.github.io}{https://guided-polynomial-diffusion.github.io}}}

\begin{document}


\maketitle
\thispagestyle{empty}
\pagestyle{empty}


\begin{abstract}

Diffusion-based motion planners are becoming popular due to their well-established performance improvements, stemming from sample diversity and the ease of incorporating new constraints directly during inference. However, a primary limitation of the diffusion process is the requirement for a substantial number of denoising steps, especially when the denoising process is coupled with gradient-based guidance. In this paper, we introduce, for the first time, diffusion in the parametric space of trajectories, where the parameters are represented as Bernstein coefficients. We show that this representation greatly improves the effectiveness of the cost-function guidance and the inference speed. We also introduce a novel stitching algorithm that leverages the diversity in diffusion-generated trajectories to produce collision-free trajectories with just a single cost function-guided model. We demonstrate that our approaches outperform current SOTA diffusion-based motion planners for manipulators and provide an ablation study on key components.

\end{abstract}


\section{INTRODUCTION}

Diffusion-based generative learning frameworks are quickly finding their ground in robotic settings. They have been explored across various domains including task planning \cite{tamp_yang2023planning, tamp_yang2023diffusion, tamp_chang2023denoising, tamp_fang2023dimsam, tamp_structdiffusion2023}, robot learning \cite{janner2022planning, robot_learning_li2023crossway}, pose estimation \cite{pose_simeonov2023rpdiff, pose_suresh2022midastouch}, grasping \cite{grasping_urain2023se, grasping_xu2023unidexgrasp}, and so on. More recently, they have been adapted for motion planning and collision avoidance in manipulators \cite{carvalho2023motion, saha2023edmp, luo2024potential}, serving as a bridge between classical and learning-based motion planners.

Classical motion planning approaches \cite{5152817_ratliff, bhardwaj2022storm, inproceedings_schulman_ho_jonathan_lee, 4082128, inproceedings, conf/nips/LikhachevGT03, Kuffner2000RRTconnectAE, LaValle1998RapidlyexploringRT, inproceedings_bangura, 7029990_erez, Williams2015ModelPP, 7487277, vamp} possess the capability to generalize across various scene types and are prevalent in industrial applications. However, optimization-based approaches rely heavily on the quality of initialization and are prone to getting trapped in local minima. On the contrary, sampling-based planners, which are known to be probabilistically complete, still require extensive hyperparameter tuning to improve their performance under a given time limit and for a given scene type \cite{9636651_lydia}. Furthermore, the trajectories generated are not smooth and therefore require additional optimization. Recently, there has been an exploration of learning-based methods as well. In these approaches, a policy is trained using a dataset of scenes and corresponding optimal solutions. This policy is either employed to assist a classical planner \cite{8793889_qureshi} or directly predict the optimal trajectory end-to-end \cite{fishman2023motion}. Such methods exhibit speed and high success rates within their training domain, without the need for extensive hyperparameter tuning. However, they fail to generalize to the out-of-distribution domains. 

Diffusion-based planners \cite{carvalho2023motion, saha2023edmp, luo2024potential} were introduced to produce success rates comparable to learning-based approaches while maintaining the generalizability of classical approaches. \cite{carvalho2023motion} introduced cost function guided diffusion for motion planning. Building further, \cite{saha2023edmp} showed that a single cost function is not sufficient to generalize across various settings. Thus, they introduce a framework with an ensemble of cost functions to improve performance. However, this requires a significant number of denoising steps, slowing down the inference speed. \cite{10610013_dipper} adapts diffusion for motion planning in quadrupeds, while \cite{yang2024diffusion} expands the capabilities of diffusion-based planners by introducing gradient-free optimization techniques for black box constraints, such as language commands. \cite{luo2024potential} proposes to train diffusion models to capture and learn each motion planning constraint separately, and compose them directly during inference. While it is shown to be successful in simple environments, it is yet to be scaled up to complex scenes, where it necessitates the composition of a large number of such models during inference. 






\begin{figure}[t!]
    \centering
    \includegraphics[width=\linewidth]{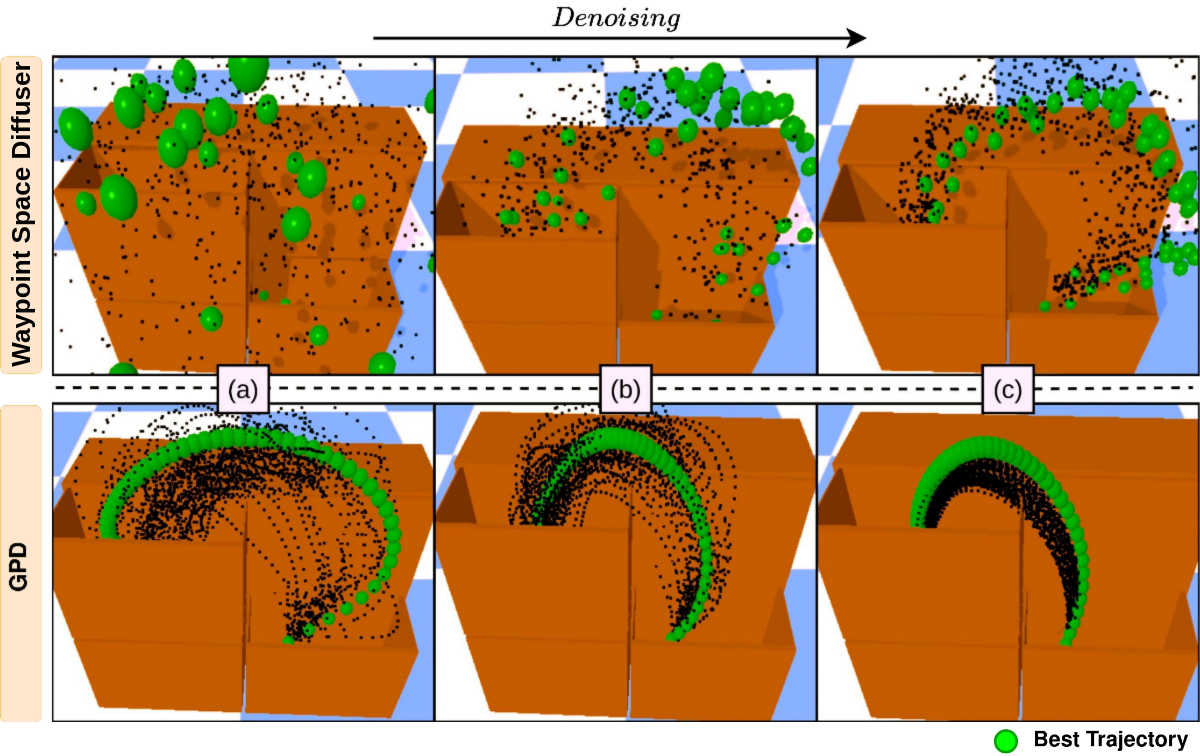}
    \caption{\small{\textbf{G}uided \textbf{P}olynomial \textbf{D}iffusion (GPD) denoises in the Bernstein space and produces smooth trajectories that converge rapidly to the prior (2nd row). In contrast, waypoint-diffusion models take a much longer time to produce trajectories that resemble the prior, as shown in the first row. (a) initial diffusion steps where GPD maintains smoothness while waypoint diffusers don't; (b) shows intermediate steps of diffusion where GPD already resembles the prior; (c) denotes the final diffusion steps.  }}
    \vspace{-1em}
    
  \label{fig:denoising}
    \vspace{-10px}
\end{figure}

One of the fundamental bottlenecks of diffusion-based motion planning is the computation time required for the denoising process. \cite{lu2022dpm, song2020denoising} attempt to increase the speed of the denoising process by skipping denoising steps and modifying the denoising process to directly estimate the target distributions. However, when denoising steps are coupled with guidance-based correction during inference, it may not be possible to arbitrarily reduce the denoising steps without compromising solution quality.

In this paper, we argue that one of the core reasons for the slow denoising process is the choice of waypoint parameterization of the trajectory used in existing works \cite{carvalho2023motion}, \cite{saha2023edmp}. Such a representation produces non-smooth trajectories for most of the denoising process. Moreover, the vanilla gradient-based guidance updates are more likely to result in non-smooth updates with waypoint parameterization. Thus, as a potential workaround, we propose Guided Polynomial Diffusion (GPD), representing trajectories as Bernstein polynomials and designing a diffusion model in polynomial coefficient space along with a gradient update rule for cost-function guidance. As a result, the denoising process is also constrained to lie in the space of continuous and differentiable polynomials, producing highly smooth trajectories within a handful of denoising steps (see Fig.\ref{fig:denoising}). We also show that such representation enhances the gradient-based guidance by ensuring smooth updates. Furthermore, the representation of a long horizon trajectory with just a low dimensional vector of polynomial coefficients allows us to fit smaller diffusion models, which further improves the inference time. 

In addition, we propose a novel stitching-based algorithm that leverages the diversity in diffusion-generated trajectories by assembling optimal segments from different trajectories directly during inference. Through empirical assessment, we demonstrate that this approach achieves superior success rates using just a single-cost function-guided diffusion model, outperforming prior methods that depend on multiple cost functions. In summary, our key contributions are:


\begin{enumerate}
    \item We propose GPD, Guided Polynomial Diffusion for motion planning which learns a prior over trajectories parameterized as Bernstein polynomials. We show that cost function guidance in this space yields significantly higher success rates with fewer diffusion steps compared to prior approaches, as shown in Table \ref{tab:guidance}, \ref{tab:main}.
    \item The Bernstein space requires a smaller diffusion model and a lower number of denoising steps to generate trajectories that resemble the prior (training dataset), leading to superior inference speeds Table \ref{tab:guidance}, \ref{tab:main}. 
    \item Additionally, we introduce a stitching algorithm, that leverages the diffusion model's ability to produce diverse trajectories. Employing just a single cost function guided diffusion model, this mechanism outperforms SOTA diffusion planners as shown in Table \ref{tab:main}.
    \item In addition to robotic manipulation, we show that the enhanced inference speeds enable application in reactive scenarios like autonomous driving and mobile robot navigation, as shown in Sec \ref{AD_results}
    
\end{enumerate}

\section{PROBLEM FORMULATION}


Consider the problem of motion planning for a robotic manipulator with $m$ joints placed in an environment $E$. We represent a trajectory $\boldsymbol{\tau}$ as a sequence of joint configurations $\boldsymbol{s}_i \in \mathbb{R}^m$ over a horizon $h$. Given an initial configuration $\boldsymbol{s}_0$ and a final configuration $\boldsymbol{s}_{h-1}$ for the trajectory $\boldsymbol{\tau} = [\boldsymbol{s}_0,\space \boldsymbol{s}_1, \space \cdots , \space \boldsymbol{s}_{h-1}] \in \mathbb{R}^{m\times h}$, we intend to minimize the cost function $J$:
\vspace{-8px}
\begin{align}
    \min_{\boldsymbol{\tau}} J(\boldsymbol{\tau}, E)
    \label{eqn:cost}
\end{align}

The cost is a function of both the trajectory $\boldsymbol{\tau}$ and the environment $E$. It takes into account the kinematic feasibility of the trajectory, self-collisions within the manipulator, and the collisions between the manipulator and the scene.











\section{DIFFUSING OVER POLYNOMIALS}
\begin{figure*}[t!]
    \centering
    \includegraphics[width=0.9\linewidth]{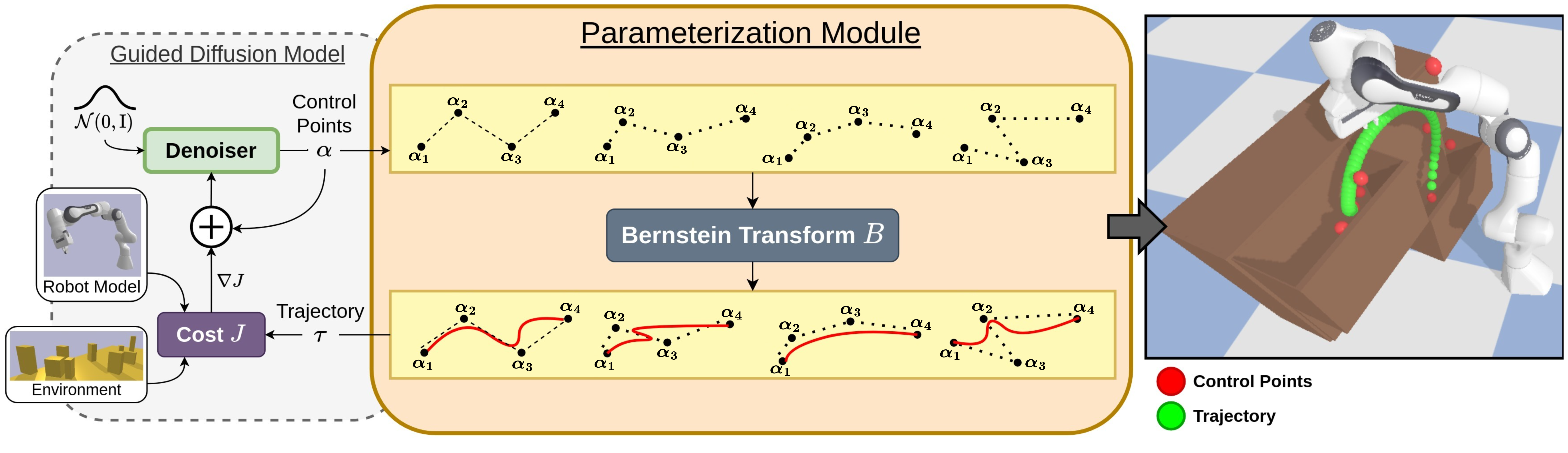}
    \caption{\small \textbf{Architecture:} GPD uses a parameterization module alongside a guided diffusion model. The diffusion model denoises the control points $\boldsymbol{\alpha}$ sampled from a Gaussian distribution and is guided by a cost function $J$. 
    The parameterization module uses a Bernstein transform to convert a set of polynomial coefficients or control points to a trajectory in the waypoint space to compute the cost function, which is used to guide the denoising process. The right-most image illustrates the control points and trajectory generated by GPD.}
    \label{fig:architecture}
    \vspace{-10px}
\end{figure*}

\begin{figure}[t]
    \centering
    \includegraphics[width=1\linewidth]{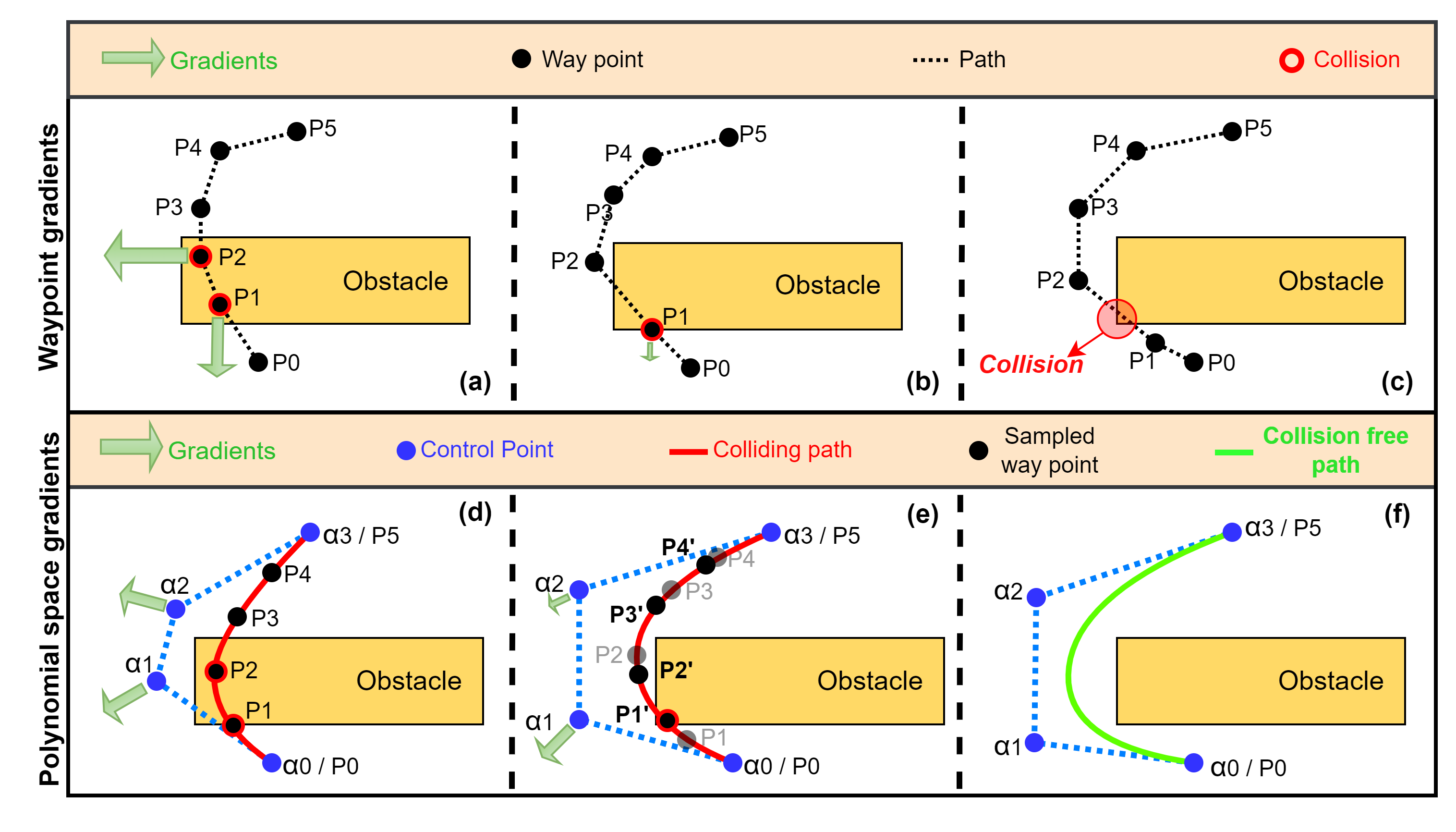}
    \caption{\small \textbf{Guidance:} GPD exhibits more effective guidance as compared to waypoint space diffusion. Row 1 shows guidance applied directly to the waypoints, where we see that just moving the individual waypoints out of collision might not lead to a final collision-free trajectory. Row 2 shows the enhanced effect of guidance in GPD for the same cost function, where the entire "string" of trajectory is moved out of collision, even when a single waypoint is in collision.
    }
    \label{fig:guidance}
    \vspace{-10px}
\end{figure}

\label{sec:diff_poly}

\noindent We parameterize our trajectory $\boldsymbol\tau$ in terms of a Bernstein polynomial on the variable $x$. This constrains the trajectory $\boldsymbol\tau$ to be a linear combination of $(c+1)$ basis polynomials. The representation is formalized as:
\begin{equation}
    \boldsymbol{\tau}(x) = \sum_{j=0}^{c}\boldsymbol{b}_j\binom{c}{j}x^j(1-x)^{c-j}
    \label{eqn:Bernstein}
\end{equation}
Each Bernstein coefficient denoted as $\boldsymbol{b}_j \in \mathbb{R}^m$ is a control point for the trajectory $\boldsymbol{\tau}$ and exists in the manipulator joint space. The variable $\tau(x)$ denotes a waypoint in the trajectory, where $x \in [0, 1]$. For the $j^{th}$ waypoint in the horizon $H$, the value of $x$ would be $\left( \frac{j}{H-1} \right)$. We compress a set of Bernstein coefficients into a vector representation denoted by $\boldsymbol{\alpha} = \left[ \boldsymbol b_0,\space \boldsymbol b_1, \space \cdots, \space \ \boldsymbol b_{c} \right] \in \mathbb{R}^{m\times (c+1)}$.

For brevity, we use a Bernstein transform $\boldsymbol{B} \in \mathbb{R}^{(c+1)\times H}$ to convert a set of coefficients $\boldsymbol{\alpha}$ to a trajectory $\boldsymbol\tau$. Given the horizon $H$, we compress the operation shown in Equation~\ref{eqn:Bernstein} to the following matrix dot product:
\vspace{-4px}
\begin{equation}
    \boldsymbol{\tau} = \boldsymbol{\alpha} \cdot \boldsymbol{B}
    \label{eqn:Bernstein_transform}
\vspace{-4px}
\end{equation}
The transform $\boldsymbol{B}$ compresses the large state-space of a trajectory $\boldsymbol\tau$ to a smaller state space of control points $\boldsymbol\alpha$. Any set of control points within the robot's joint limits represents a smooth trajectory. Therefore, the transform maintains the temporal consistency and smoothness associated with a trajectory. Bernstein parameterization establishes limits on the trajectory by confining it within the convex hull defined by the Bernstein control points. 




\subsection{Diffusion Models}

\noindent Diffusion models \cite{sohl2015deep, ho2020denoising} fall within a group of generative models where Gaussian noise is sequentially added to a dataset sample through a scheduled forward diffusion process. This transforms the sample into an isotropic Gaussian. We denote the forward process on the control points $\boldsymbol\alpha$ as $q(\boldsymbol{\alpha}_t|\boldsymbol{\alpha}_{t-1}) = \mathcal{N}(\boldsymbol{\alpha}_t; \sqrt{1 - \boldsymbol{\beta}_t}\boldsymbol{\alpha}_{t-1}, \boldsymbol{\beta}_t \text{I})$ where $\boldsymbol{\beta}_t$ is the variance schedule and $t$ is the diffusion timestep. 
A trainable reverse diffusion process $p_{\theta}(\boldsymbol{\alpha}_{t-1}/\boldsymbol{\alpha}_t) = \mathcal{N}(\boldsymbol{\alpha}_{t-1}; \mu_{\theta}(\boldsymbol{\alpha}_t, t), \boldsymbol{\Sigma}_t)$ parametrized by $\theta$ transforms a sample $\boldsymbol{\alpha}_T \sim \mathcal{N}(0, \text{I})$ from a standard normal distribution into a valid sample from the prior $p_{\theta}(\boldsymbol{\alpha})$, by sequentially removing Gaussian noise.


We train a diffusion model to learn the prior $p_\theta(\boldsymbol{\alpha})$ over a set of Bernstein polynomial coefficients $\boldsymbol\alpha$. During inference, we repeatedly fix the first and the last control point at every denoising step, to condition the diffusion model to produce a trajectory for the given start and goal control points. These start and goal control points directly correspond to the start and goal configurations in the joint space.

\begin{algorithm}[t]
\DontPrintSemicolon
\caption{Guided Polynomial Diffusion}
\KwInput{Learned reverse diffusion ${\mu}_{\theta}(\boldsymbol{\alpha}_t, t)$, Covariance schedule $\boldsymbol{\Sigma}_t$, $n$ guide cost functions $J$, learning rate schedule $\boldsymbol\lambda$, hyper-parameter schedule $\mathbf{o}_t$}
\KwInitialization{$\boldsymbol{\alpha}_T \sim \mathcal{N} (\textbf{0}, \textbf{I})$}
\For{$t = T, \dots, \dots 1$}{
     \textcolor{gray}{\small\texttt{\# Compute gradients wrt coefficients for each guide in parallel }} \\
    $^{1 \dots n}\boldsymbol q_t = \boldsymbol B \cdot ^{1 \dots n} \boldsymbol \alpha_t$ \\
     $^{1 \dots n}{\boldsymbol{\mathcal{G}}} = \boldsymbol B^T \cdot \nabla_{^{1 \dots n}\boldsymbol q_t}(J(^{1 \dots n}\boldsymbol q_t, ^{1 \dots n}\boldsymbol o_t))$ \\
     \textcolor{gray}{\small\texttt{\# Denoise for n guides in parallel}} \\ 
   ${^{1 \dots n}}\boldsymbol{\alpha_{t-1}} \sim \mathcal{N}(\boldsymbol{\mu}_{\theta}( {^{1 \dots n}} \boldsymbol{\alpha}_t, t )+{^{1 \dots n}}\lambda_t \cdot ^{1 \dots n}{\boldsymbol{\mathcal{G}}}, ^{1 \dots n}\boldsymbol\Sigma_t)$ 
   }
\textbf{Return} $\boldsymbol{\tau}_0 = \boldsymbol B\cdot {^j\boldsymbol \alpha_0}$ corresponding to minimum collision cost.
\label{algo:gpd}
\end{algorithm}

\subsection{Guidance on Coefficients}
\label{sec:guidance}



\noindent The denoising process can generate trajectory samples from a distribution that models the dataset but does not account for scene-specific collision costs. To produce trajectories that comply with the scene-specific collision constraints, we guide the diffusion process with the gradients of the cost function $J$ with respect to the control points i.e. Bernstein coefficients $\boldsymbol\alpha$, following
\begin{align}
     \boldsymbol\alpha_{t-1} = \boldsymbol\alpha_{t} - \gamma \epsilon - \gamma_2 \nabla_{\boldsymbol\alpha_t} J(\boldsymbol\alpha_{t}, \boldsymbol o_t)  + \boldsymbol\zeta
\end{align}
where $\boldsymbol o_t$ are the cost function parameters, and $\boldsymbol\zeta \sim N(0, \boldsymbol \Sigma_{t}) $.
To compute $\nabla_{\boldsymbol \alpha_t} J(\boldsymbol \alpha_{t}, \boldsymbol o_t)$, we first discretize the polynomial using the equation $\boldsymbol q_t = \boldsymbol \alpha_{t} \cdot \boldsymbol B$, where the sampled waypoints of the $\boldsymbol q_t$ are equidistant from each other.
We then compute $J(\boldsymbol q_t, \boldsymbol o_t)$ using the differentiable collision cost functions, similar to \cite{saha2023edmp}. Finally, using chain rule of differentiation, we compute $\nabla_{\boldsymbol \alpha_t} J(\boldsymbol \alpha_{t}, \boldsymbol o_t)$ as:
\begin{align}
    \nonumber \nabla_{\boldsymbol \alpha_t}J(\boldsymbol q_{t}, \boldsymbol o_t) &= \nabla_{\boldsymbol \alpha_t}(\boldsymbol q_t) \nabla_{\boldsymbol q_t}(J(\boldsymbol q_t, \boldsymbol o_t)) \\ &= \boldsymbol B^T \cdot \nabla_{\boldsymbol q_t}(J(\boldsymbol q_t, \boldsymbol o_t))
    \label{grad_precond_gpd}
\end{align}

 The R.H.S of \eqref{grad_precond_gpd} can also be understood as pre-conditioning the gradient obtained by waypoint parameterization with the matrix $\boldsymbol B ^T$. This has connections to the idea presented in \cite{5152817_ratliff}, which uses a finite-differences matrix to pre-condition cost gradients obtained for waypoint parameterized trajectories. However,  $\boldsymbol B ^T$ is a better pre-conditioner since it ensures higher continuity and differentiability in the updates. This is similar to pulling a string, where gradients from each waypoint influence all coefficients, shifting the entire trajectory while maintaining smoothness. This contrasts with updating the waypoints directly with their corresponding gradients, where each waypoint moves independently of each other (Fig. \ref{fig:guidance}). We believe this is the first use of such preconditioning in diffusion model guidance updates. 
 
 Furthermore, we argue that cost function guidance in GPD is more effective as it applies gradients to smoother, prior-resembling trajectories earlier in denoising, as seen in Fig. ~\ref{fig:denoising}. This contrasts with prior guided diffusion methods, which apply gradients to non-smooth waypoints in most of the intermediate denosing steps. Such methods are less effective as guidance on noisy trajectories is often dominated by the denoising process pushing the trajectories towards the prior, rather than towards optimising the cost.

\begin{algorithm}[t]
\DontPrintSemicolon
\caption{GPD with Stitching}
\label{algo:stitching}
\KwInput{Trajectories generated by GPD $\boldsymbol{\mathcal{T}}_{gpd}$, Goal $G$, Cost function $J$, Collision window size $w$
}
\KwInitialization{$i={arg\,min}$ \space $J(\boldsymbol{\mathcal{T}}_{gpd})$, $\boldsymbol \tau_{sol}=$ [ ], $j = 0$
}

\While{$\tau_i^j \notin G$}{
    $\boldsymbol C_{free} = collision\_fn(\boldsymbol \tau_i^{j:j + w}, E)$
    
    \If{$! C_{free}$}{
        $i_{stitched}$, $j_{stitched}$, $\boldsymbol\tau_{stitch}$ = \textbf{get_valid_stitch}($\boldsymbol{\mathcal{T}}_{gpd}$, $\tau_i^j$)

        $\boldsymbol\tau_{sol}.add(\boldsymbol\tau_{stitch})$

        i, j = $i_{stitched}$, $j_{stitched}$\\
        
    }

    $\boldsymbol\tau_{sol}.add(\boldsymbol\tau_i^j)$ \\
}

\textbf{Return} $\boldsymbol\tau_{sol}$
\end{algorithm}

\begin{figure}
    \centering
    \includegraphics[width=1\linewidth]{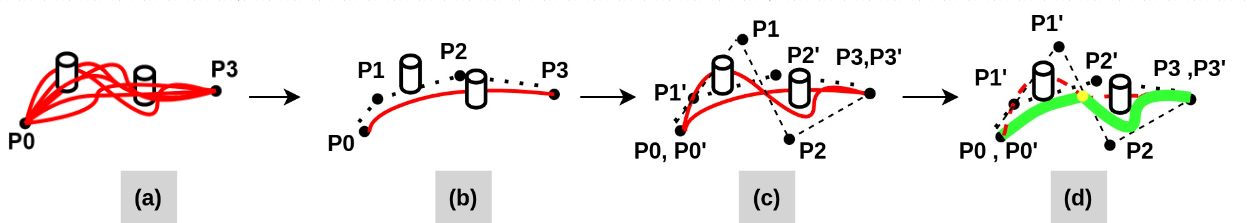}
    \caption{\small \textbf{Stitching:} (a) Diverse batch of generated trajectories. (b) trajectory with the lowest cost. (c) pair of trajectories that can be stitched. (d) shows the final collision-free trajectory by stitching two different trajectories. The stitch is shown in yellow.}
    \label{fig:stitching}
    \vspace{-10px}
\end{figure}
\subsection{Stitching}
\label{sec:stitching}






Prior work \cite{saha2023edmp} shows the need for an ensemble of cost-guided models, however, this would require fine hyper-parameter tuning of multiple cost functions which can be a tedious process. We circumvent this problem by leveraging the diversity of generated trajectories. Specifically, we observe that various trajectories traverse through distinct collision-free segments as shown in Fig. \ref{fig:stitching}, which when stitched together can produce fully collision-free trajectories.




Let $E$ represent the scene as a collection of obstacles. Let $\boldsymbol{\mathcal{T}}_{gpd} =\left[ \boldsymbol\tau_1, \boldsymbol\tau_2, ... , \boldsymbol\tau_n \right]$ be a batch of $n$ trajectories generated by GPD, and $\tau_{i}^{j}$ be the $j^{th}$ waypoint in trajectory $\boldsymbol\tau_{i}$. Let $E_i^{free} \subseteq E$ denote segments in the scene that do not collide with the robot in trajectory $\boldsymbol\tau_i$. We suggest that when a set of trajectories $\boldsymbol{\mathcal{T}}$ is highly diverse, $\cup_i E_i^{free} = E$, even in cases where $E_i^{free} \neq E \ \forall \ i\in \{1, 2, ... , n\}$. This implies that as we increase the diversity, we are more likely to find a set of collision-free segments that can be stitched together to obtain a fully collision-free trajectory. To leverage this property, we propose a stitching framework, that merges trajectories during inference. 

Initially, we rollout the lowest-cost trajectory $\boldsymbol\tau_i$ from $\boldsymbol{\mathcal{T}}_{gpd}$ using a sliding window approach. We denote $e_i^{free} $ as the segments in the window that do not collide with the robot. During the rollout, we sequentially evaluate if $e_i^{free} \neq E$, in which case we look for a valid stitch target $\boldsymbol\tau_{s}$ in $\boldsymbol{\mathcal{T}}_{gpd}$. We denote a stitch target $\tau_{s}$ as valid, if the closest waypoint $\tau_{s}^{p}$ in $\tau_{s}$ (in the direction of goal) is collision-free in its respective sliding window. Given a valid stitch target, we use RRT-C~\cite{844730} as a local planner to obtain a stitch $\tau_{stitch}$. The process is continued until the goal configuration is reached. Algorithm~\ref{algo:stitching} summarizes our stitching framework. 

In polynomial diffusion, the intermediate diffusion steps yield smooth trajectories that converge rapidly to our kinematically valid prior. By the final diffusion steps, the trajectories are both smooth and kinematically valid, closely resembling the prior (as shown in Fig. \ref{fig:denoising}) while also incorporating some applied guidance. Hence including the trajectories obtained from the last few diffusion steps results in a batch of more diverse trajectories, due to the combination of trajectories at various stages of guidance and denoising. Our results demonstrate that even with a single guide, we can achieve higher success rates. To test the diversity of our prior, we benchmark the performance of our stitching algorithm in Section~\ref{sec:stitching_results}.


\section{EXPERIMENTAL SETUP}
\label{sec:experiments}


\begin{table}[t]
\begin{center}
\adjustbox{max width=\linewidth}{
\centering
\begin{tabular}{c|ccc|c}
\Xhline{3\arrayrulewidth}
& \multicolumn{3}{c|}{Per Dataset Success (\%, $\uparrow$)} & Planning \\
Method & Global & Hybrid & Both & Time (s, $\downarrow$) \\
\Xhline{2\arrayrulewidth}
PD & 32.11 & 40.38 & 41.54 & \textbf{0.22} \\
MPD \cite{carvalho2023motion} & 51.33	 & 62.27 & 67.12 & 7.32\\
EDMP-1G \cite{saha2023edmp} & 53.11 & 66.50 & 68.88 & 7.95 \\
\textbf{GPD-1G} & \textbf{65.50} & \textbf{72.89} & \textbf{73.00} & 0.8 \\
\Xhline{3\arrayrulewidth}
\end{tabular}}
\end{center}
\vspace{-8px}
\captionsetup{textfont=normalfont}
\caption{\small Comparison of GPD-1G against PD and guidance with a single cost function in the waypoint space (EDMP-1G, MPD), in terms of success rates and inference time.
}
\label{tab:guidance}
\vspace{-8px}
\end{table}


We conduct our experiments using the Franka Panda robot
in the Pybullet simulator \cite{coumans2021}.

\textbf{Training}: We train our model on a dataset of 6.54 million Bernstein polynomials. The polynomials are generated by fitting their coefficients using least squares to the M$\pi$Nets~\cite{fishman2023motion} training dataset, comprising of 6.54 million trajectories
.We use 8th-order Bernstein polynomials with 8 control points to parameterize the train set trajectories, as they provide a sufficient fit.
The M$\pi$Nets dataset consists of tabletops, dressers, and cupboards out of which 3.27 million trajectories generated using a global planner, AIT* \cite{strub2020adaptively}, and another 3.27 million trajectories using a hybrid planner, that uses AIT* \cite{strub2020adaptively} to generate plans in the end-effector space, and Geometric Fabrics \cite{xie2021geometric} to produce geometrically consistent motion in joint space. We model our polynomial denoiser as a temporal UNet similar to \cite{janner2022planning}, and train it for 20k epochs on a RTX A4000 GPU. We set the number of diffusion timesteps $T = 64$, in comparison to EDMP~\cite{saha2023edmp} which uses $T= 256$.



\textbf{Testing}: We benchmark our performance on 3 different test datasets from M$\pi$Nets~\cite{fishman2023motion}. (1) A \textit{Global Solvable Dataset} consists of 1800 environments on which the \textit{global planner} was able to generate valid collision-free trajectories. (2) A \textit{Hybrid Solvable Dataset} consists of 1800 environments on which only the \textit{hybrid planner} was able to generate valid collision-free trajectories. (3) A \textit{Both Solvable Dataset} consists of 1800 environments where both planners (1 \& 2) were able to generate valid collision-free trajectories.
Each scene is defined by an obstacle configuration, an initial joint configuration and a goal pose. We perform our benchmarks on a workstation with a Ryzen 9 5950X processor and an RTX 3060 GPU.



\begin{figure}[t!]
    \centering
    \includegraphics[width=0.7\linewidth]{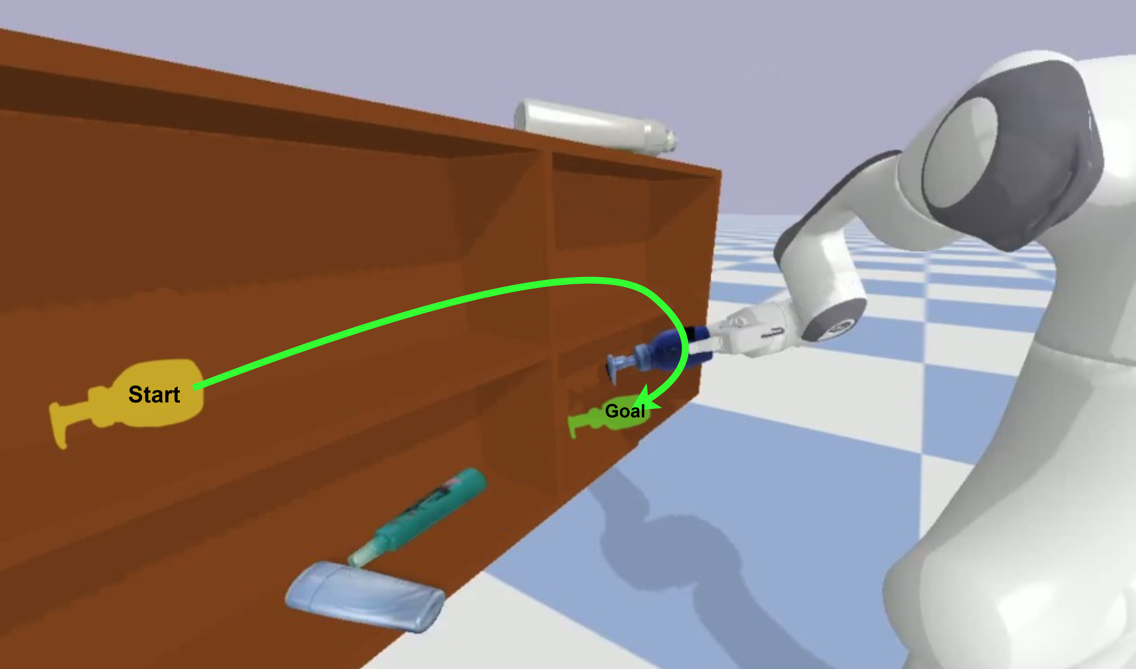}
    \caption{\small \textbf{Out-of-distribution scenes:} GPD generalises to unseen object-in-hand scene \textbf{without any additional training} just by modifying our cost function directly at inference}
    \label{fig:obj_in_hand}
    \vspace{-3px}
\end{figure}

\begin{figure}[t!]
    \centering
    \includegraphics[width=0.85\linewidth]{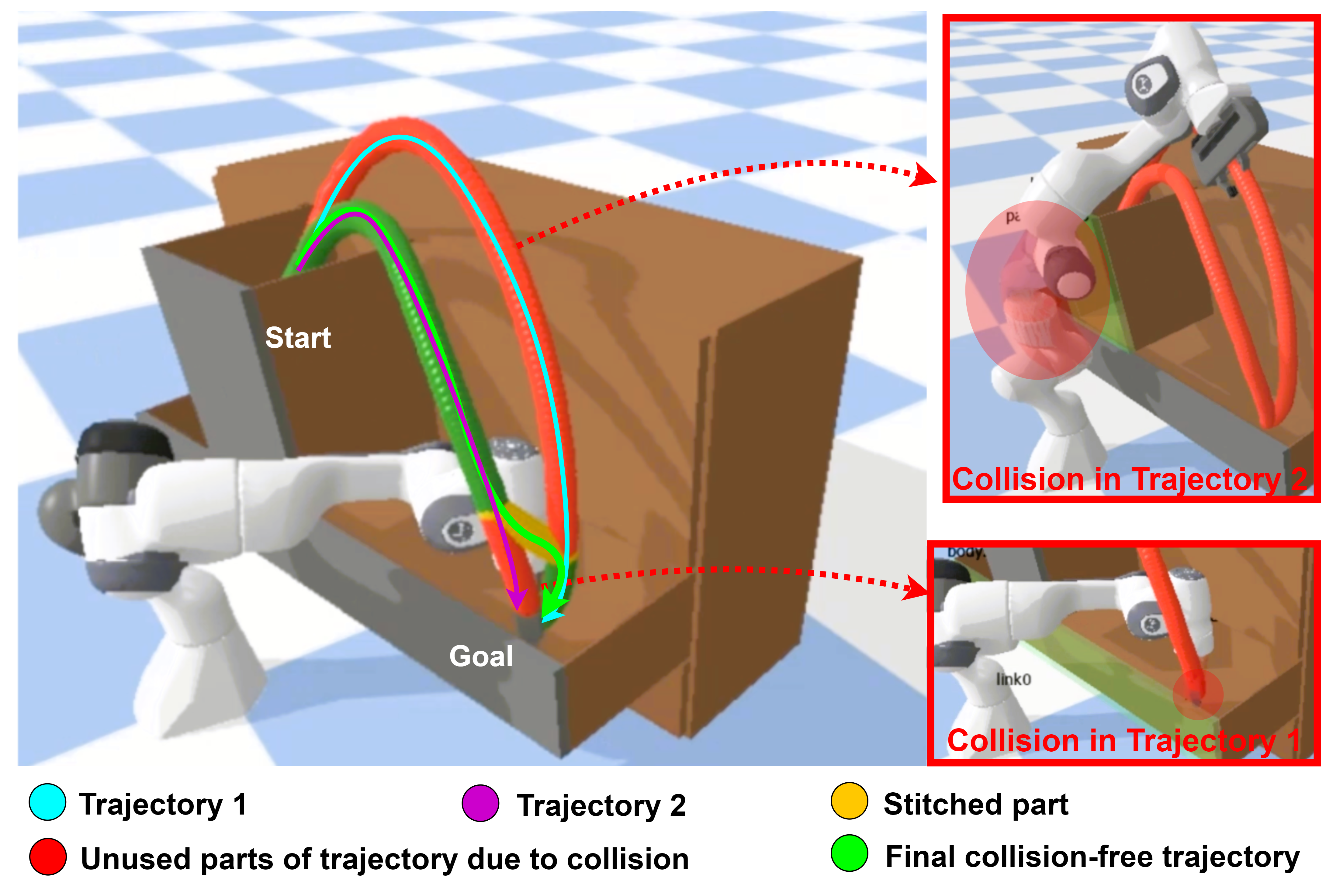}
    \caption{\small \textbf{Qualitative Results of Stitching.} Collision is detected in the red portion of the initial selected trajectory. The stitching algorithm then generates a stitch (shown in yellow) to another trajectory (which is in collision at a different segment). The final executed trajectory is shown in green.}
    \label{fig:stitching_quali}
    \vspace{-1em}
\end{figure}


\textbf{Metrics}: We use Success Rate (SR) as the main metric to compare our performance against baselines. The success rate is evaluated to be the percentage(\%) of successful runs in the test dataset. If the robot end-effector successfully reaches the goal pose, without colliding with the environment or itself, we report a score of 1, otherwise 0. Collisions are checked using Pybullet's in-built collision checker, by executing the planned trajectory in the bullet physics engine. We also benchmark the planning time of our algorithms, and stitching time for algorithms that use the stitching method. 

\textbf{Models and Baselines}: We benchmark the performances of six different models that use polynomial parameterization. 
(1) GPD-1G - GPD with a single cost function-based guidance, (2) GPD-7G - GPD with 7 cost functions with varying configurations similar to ~\cite{saha2023edmp} (3) PD - Polynomial diffusion prior, (4) GS - Stitching applied to Bernstein polynomials sampled from a Gaussian distribution, (5) PDS - Stitching applied to trajectories generated by PD, (6) GPDS - Stitching applied to trajectories generated by GPD-1G. We compare our model against stochastic optimization-based planners (G.Fabrics~\cite{xie2021geometric}, STORM~\cite{bhardwaj2022storm}), optimization-based planner CHOMP~\cite{5152817_ratliff} with a batch of quintic-spline initializations, and behavior-cloning based deep-learning planners (MPNets~\cite{8793889_qureshi},  M$\pi$Nets~\cite{fishman2023motion}) trained on the same dataset.The scores for G. Fabrics, STORM and M$pi$nets, have been reported as per the ones mentioned in the M$pi$nets paper. We also compare against guided-diffusion based models MPD~\cite{carvalho2023motion} and EDMP~\cite{saha2023edmp}. We adapted MPD to the more diverse M$\pi$Nets environment and for a fair comparison MPD is trained with different number diffusion steps and UNet sizes including the ones mentioned in their paper, and the best scores have been reported. Similarly, EDMP scores are reported with the best 12 Guides provided by them.


\subsection{Effect of Parameterization on Guidance}
\label{sec:guidance_results}
We analyse the effect of parameterizing trajectories with Bernstein polynomial coefficients on Guidance by benchmarking the GPD-1G against PD, and also with SOTA waypoint diffusion models EDMP \cite{saha2023edmp} and MPD \cite{carvalho2023motion}. Table~\ref{tab:guidance} shows that while unguided polynomial prior (PD) is able to achieve good success rates, scene-specific cost guided frameworks (MPD, EDMP, GPD) significantly improve the success rates. We further find that our GPD-1G clearly outperformes both EDMP-1G and MPD in terms of both success rate, and speed. This increase in success rate is attributed to the increased effect of guidance as mentioned in Section \ref{sec:guidance}. Furthermore, GPD is about 10 times faster than MPD and EDMP-1G, taking only 0.8s to produce collision-free trajectories. This is due to the polynomial parameterization, which enhances convergence speed, reduces the size of the state space, and decreases the complexity of the diffusion model. Note that MPD used on these scenes has a higher number of diffusion steps as compared to the one mentioned in ~\cite{carvalho2023motion}; this is because of the increase in diversity and the complexity of the prior trajectories in the M$\pi$Nets scenes.



\subsubsection*{\textbf{Comparision with baselines}}
We first evaluate our GPD-7G model, with 3 cost functions based on the intersection volume and 4 cost functions based on the swept volume costs similar to ~\cite{saha2023edmp} on the test datasets. We show SOTA peformance when compared to prior diffusion based methods~\cite{carvalho2023motion, saha2023edmp} and classical methods \cite{5152817_ratliff, xie2021geometric, bhardwaj2022storm}, while using lower number of cost functions than the previous SOTA \cite{saha2023edmp}. We also compare with the learning based approaches M$\pi$Nets and MPNets, which are expected to show high success rates as they are trained specifically on these datasets. Our method achieves SOTA performance in the global dataset and demonstrates comparative performance to the SOTA planner (M$\pi$Nets) in Hybrid and Both datasets, despite being trained in a scene-agnostic manner. 
Additionally, similar to \cite{carvalho2023motion, saha2023edmp}, we generalize to out-of-distribution settings, such as Object-in-hand (Fig.~\ref{fig:obj_in_hand}) by incorporating the object as another link of our manipulator with fixed joints and adding an object-environment collision-cost to our cost function directly during inference. Thus, we generalize to new settings without needing any additional training, unlike  M$\pi$Nets and MPNets.

\begin{table}[t]
\begin{center}
\adjustbox{max width=\linewidth}{
\centering
\begin{tabular}{{c |@{\hspace{5pt}}ccc@{\hspace{2pt}} | c}}
\Xhline{3\arrayrulewidth}
\multirow{2}{*}{Method} & \multicolumn{3}{c|}{Per Dataset Success (\%, $\uparrow$) } & \multirow{2}{*}{Time (s, $\downarrow$)} \\
& Global & Hybrid & Both & \\
\Xhline{2\arrayrulewidth} 
G. Fabrics ~\cite{xie2021geometric} & 38.44 & 59.33 & 60.06 & 0.15 \\ 
STORM ~\cite{bhardwaj2022storm} & 50.22 & 74.50 & 76.00 & 4.03 
\\ \Xhline{1\arrayrulewidth} 
CHOMP ~\cite{5152817_ratliff} & 26.67 & 31.61 & 32.2 & 2.39
\\ \Xhline{1\arrayrulewidth}
MPNets ~\cite{8793889_qureshi} & 41.33 & 65.28 & 67.67 & 4.95 \\ 
M$\pi$Nets ~\cite{fishman2023motion} & 75.78 & \textbf{95.33} & \textbf{95.06} & 0.33 
\\ \Xhline{1\arrayrulewidth}
MPD ~\cite{carvalho2023motion} & 46.33	 & 57.27 & 62.12 & 7.32 \\ 
EDMP ~\cite{saha2023edmp} & 75.93 & 86.13 & 85.06 & 8.22 \\ 
\textbf{GPD 7G} & 81.94	& 90.88 & 90.24 & 0.85 
\\ \Xhline{1\arrayrulewidth}
GS & 43.31 & 57.92 & 54.37 & 7.48 ( 0 D + 7.48 S) \\
PDS & 68.81 & 80.19 & 75.49 & 4.17 (0.22 D + 3.95 S) \\
\textbf{GPDS} & \textbf{87} & 92.83 & 92.61 & 1.9 (0.8 D + 1.1 S) \\

\Xhline{3\arrayrulewidth}
\end{tabular}}
\end{center}
\vspace{-8px}
\captionsetup{textfont=normalfont}
\caption {\small Comparison of our models in terms of success rate, and planning time (in seconds) on the M$\pi$Nets~\cite{fishman2023motion} dataset. D refers to denoising time, and S is the time taken by the stitching algorithm}
\label{tab:main}
\vspace{-8px}
\end{table}




\subsection{Leveraging diversity to increase performance}
\label{sec:stitching_results}
Using an ensemble of cost functions greatly improves the success rates as shown in ~\cite{saha2023edmp}. However, each cost function needs to be carefully tuned, which makes it a tedious process. We show that our stitching algorithm, applied on a single cost function guided polynomial diffusion (GPDS), outperforms both single and multi-cost function-guided diffusion baselines \cite{carvalho2023motion, saha2023edmp} (as shown in Table \ref{tab:main}). The stitching algorithm leverages the diversity of the generated trajectories as explained in Section~\ref{sec:stitching} and illustrated in Fig. \ref{fig:stitching_quali}. We use RRT-Connect \cite{Kuffner2000RRTconnectAE} as the local planner to stitch between the generated trajectories. 

We show all the results related to this section on Table ~\ref{tab:main} where we compare stitching on GPDS, GS, and PDS.  We use a batch size of 1000 trajectories for stitching on GS. For PDS and GPDS we use a batch size of 32 during the denoising process and include trajectories from the last 5 diffusion steps. This increases the effective batch size for stitching by a factor of 5 by directly leveraging the fact that the trajectories generated by polynomial diffusion in the last few timesteps closely resemble the prior distribution, as illustrated in Fig. \ref{fig:denoising} and explained in Sec \ref{sec:stitching}.

The table shows that GPDS, which utilizes a single cost function, outperforms GPD 7G and EDMP, both of which use an ensemble of cost functions to boost their overall success rates. Next, we show that stitching requires good sample quality of generated trajectories. For this, in the same table, we first compare PDS against GS, highlighting the point that stitching on kinematically valid trajectories (PDS) far outperforms stitching on Gaussian-sampled trajectories (GS). In addition, since GPD ensures that most segments of the trajectories are collision-free through guidance, we see that stitching on such trajectories (GPDS) outperforms stitching on just kinematically feasible prior (PDS). 

We also benchmark GPDS against an optimized version of RRT-Connect, to show that GPDS's superior performance is a result of combining the local planner used for stitching (RRT-Connect) and the GPD prior. RRT-connect achieves a success rate of 83.33, 90.83, and 90.60 on the global, hybrid, and both datasets respectively. As evident from the Section \ref{tab:main}, GPDS outperforms both RRT-Connect and GPD 1G. Note that RRT-Connect takes an average time of 7.14s when benchmarked on the M$\pi$Nets dataset, whereas, we only take an average of 1.9s.

\subsection{GPD for Navigation}
\label{AD_results}
\begin{figure}[t!]
    \centering
    \includegraphics[width=0.65\linewidth]{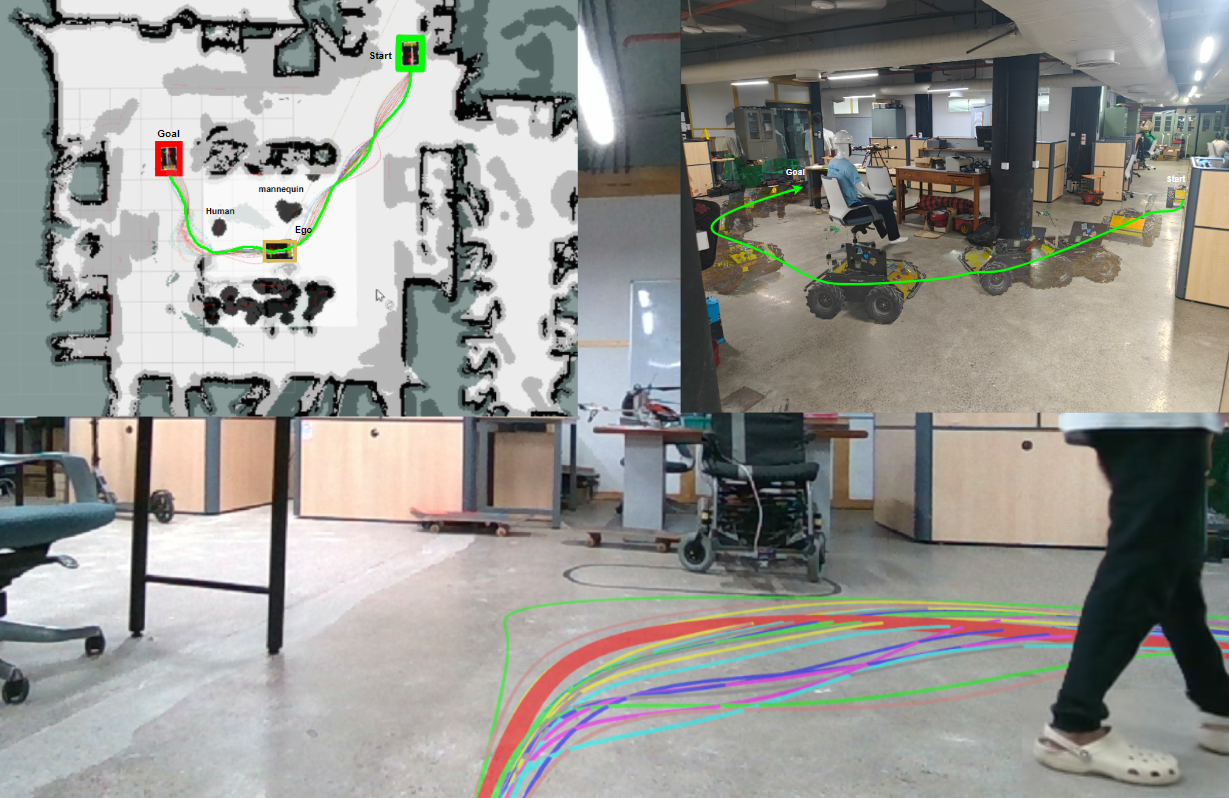}
    \caption{\small \textbf{Multi-modal trajectory generation in an indoor space.} The minimum-cost trajectory is shown in Red, along with diverse navigable paths projected onto a perspective view.}
    \label{fig:husky}
    \vspace{-3px}
\end{figure}

\begin{figure}[t!]
    \centering
    \includegraphics[width=1\linewidth]{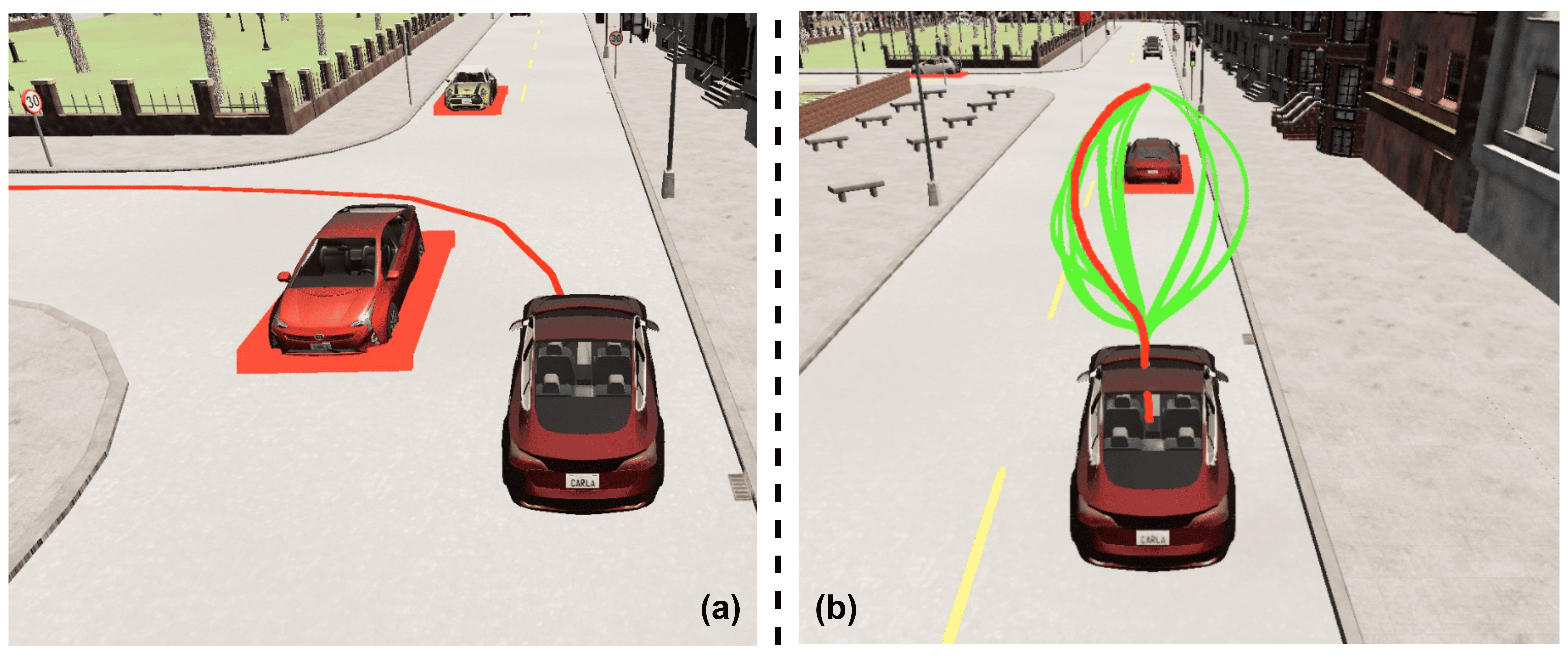}
    \caption{\small \textbf{GPD performance in a simulated urban driving scene.} (a) shows a collision-free trajectory for a turning scenario. (b) shows a diverse trajectory set generated by GPD with the selected plausible path highlighted in Red. The relevant obstacles considered for collision-checking are shown with a red patch beneath them.}
    \label{fig:overtake}
    \vspace{-1.5em}
\end{figure}

\noindent The non-limiting nature of our model is also demonstrated for time-critical requirements of navigation tasks across distinct environments. Fig. \ref{fig:husky} and \ref{fig:overtake} showcase the model's ability to generate multi-modal trajectories in complex indoor and outdoor conditions respectively, where the denoising process is guided by a cost function that penalizes curvature and acceleration magnitudes. In the urban driving scenario, shown in Fig. \ref{fig:overtake}, the presence of dynamic obstacles necessitates the inclusion of collision cost to plan collision-free trajectories at reactive rates. In both contexts, GPD was able to generate smooth, feasible trajectories at a frequency of 16 Hz, demonstrating near real-time planning capabilities due to the model's enhanced computational speed.



\section{CONCLUSION \& FUTURE WORK}

In this paper, we enhanced previous diffusion-based motion planners by learning a prior over Bernstein polynomial coefficients to represent valid trajectories, in contrast to waypoint representation. We show that cost function guidance during inference is more effective when denoising in the polynomial space rather than the waypoint space. We also introduce a stitching algorithm to leverage the diversity of generated trajectories directly during inference. Through these, we show SOTA performance at both speed and success rates. 
Finally, with the enhanced inference speed, we demonstrate our ability to adapt to reactive scenes such as in an autonomous driving setting. 

Future research could explore implicit modelling of open-vocabulary cost functions from data, potentially enabling language-directed motion planning without explicit cost functions.

\printbibliography

@article{lu2022dpm,
  title={Dpm-solver: A fast ode solver for diffusion probabilistic model sampling in around 10 steps},
  author={Lu, Cheng and Zhou, Yuhao and Bao, Fan and Chen, Jianfei and Li, Chongxuan and Zhu, Jun},
  journal={Advances in Neural Information Processing Systems},
  volume={35},
  pages={5775--5787},
  year={2022}
}

@article{song2020denoising,
  title={Denoising diffusion implicit models},
  author={Song, Jiaming and Meng, Chenlin and Ermon, Stefano},
  journal={arXiv preprint arXiv:2010.02502},
  year={2020}
}

@InProceedings{yang2024diffusion,
    author    = {Yang, Brian and Su, Huangyuan and Gkanatsios, Nikolaos and Ke, Tsung-Wei and Jain, Ayush and Schneider, Jeff and Fragkiadaki, Katerina},
    title     = {Diffusion-ES: Gradient-free Planning with Diffusion for Autonomous and Instruction-guided Driving},
    booktitle = {Proceedings of the IEEE/CVF Conference on Computer Vision and Pattern Recognition (CVPR)},
    month     = {June},
    year      = {2024},
    pages     = {15342-15353}
}

@inproceedings{
    luo2024potential,
    title={Potential Based Diffusion Motion Planning},
    author={Yunhao Luo and Chen Sun and Joshua B. Tenenbaum and Yilun Du},
    booktitle={Forty-first International Conference on Machine Learning},
    year={2024},
  }

@INPROCEEDINGS{vamp,
  author={Thomason, Wil and Kingston, Zachary and Kavraki, Lydia E.},
  booktitle={2024 IEEE International Conference on Robotics and Automation (ICRA)}, 
  title={Motions in Microseconds via Vectorized Sampling-Based Planning}, 
  year={2024},
  volume={},
  number={},
  pages={8749-8756},
  doi={10.1109/ICRA57147.2024.10611190}}

@INPROCEEDINGS{saha2023edmp,
  author={Saha, Kallol and Mandadi, Vishal and Reddy, Jayaram and Srikanth, Ajit and Agarwal, Aditya and Sen, Bipasha and Singh, Arun and Krishna, Madhava},
  booktitle={2024 IEEE International Conference on Robotics and Automation (ICRA)}, 
  title={EDMP: Ensemble-of-costs-guided Diffusion for Motion Planning}, 
  year={2024},
  volume={},
  number={},
  pages={10351-10358},
  doi={10.1109/ICRA57147.2024.10610519}}

@ARTICLE{4082128,
  author={Hart, Peter E. and Nilsson, Nils J. and Raphael, Bertram},
  journal={IEEE Transactions on Systems Science and Cybernetics}, 
  title={A Formal Basis for the Heuristic Determination of Minimum Cost Paths}, 
  year={1968},
  volume={4},
  number={2},
  pages={100-107},
  doi={10.1109/TSSC.1968.300136}}

@inproceedings{conf/nips/LikhachevGT03,
  author = {Likhachev, Maxim and Gordon, Geoffrey J. and Thrun, Sebastian},
  booktitle = {NIPS},
  crossref = {conf/nips/2003},
  editor = {Thrun, Sebastian and Saul, Lawrence K. and Schölkopf, Bernhard},
  ee = {http://papers.nips.cc/paper/2382-ara-anytime-a-with-provable-bounds-on-sub-optimality},
  isbn = {0-262-20152-6},
  pages = {767-774},
  publisher = {MIT Press},
  title = {ARA*: Anytime A* with Provable Bounds on Sub-Optimality.},
  url = {http://dblp.uni-trier.de/db/conf/nips/nips2003.html#LikhachevGT03},
  year = 2003
}

@inproceedings{inproceedings,
author = {Likhachev, Maxim and Ferguson, David and Gordon, Geoffrey and Stentz, Anthony and Thrun, Sebastian},
year = {2005},
month = {01},
pages = {262-271},
title = {Anytime Dynamic A*: An Anytime, Replanning Algorithm.},
journal = {Proceedings of the International Conference on Automated Planning and Scheduling (ICAPS)}
}

@article{LaValle1998RapidlyexploringRT,
  title={Rapidly-exploring random trees : a new tool for path planning},
  author={Steven M. LaValle},
  journal={The annual research report},
  year={1998},
  url={https://api.semanticscholar.org/CorpusID:14744621}
}

@article{Kuffner2000RRTconnectAE,
  title={RRT-connect: An efficient approach to single-query path planning},
  author={James J. Kuffner and Steven M. LaValle},
  journal={Proceedings 2000 ICRA. Millennium Conference. IEEE International Conference on Robotics and Automation. Symposia Proceedings (Cat. No.00CH37065)},
  year={2000},
  volume={2},
  pages={995-1001 vol.2},
  url={https://api.semanticscholar.org/CorpusID:17124403}
}

@article{Williams2015ModelPP,
  title={Model Predictive Path Integral Control using Covariance Variable Importance Sampling},
  author={Grady Williams and Andrew Aldrich and Evangelos A. Theodorou},
  journal={ArXiv},
  year={2015},
  volume={abs/1509.01149},
  url={https://api.semanticscholar.org/CorpusID:14146342}
}

@INPROCEEDINGS{7487277,
  author={Williams, Grady and Drews, Paul and Goldfain, Brian and Rehg, James M. and Theodorou, Evangelos A.},
  booktitle={2016 IEEE International Conference on Robotics and Automation (ICRA)}, 
  title={Aggressive driving with model predictive path integral control}, 
  year={2016},
  volume={},
  number={},
  pages={1433-1440},
  doi={10.1109/ICRA.2016.7487277}}

@inproceedings{bhardwaj2022storm,
  title={Storm: An integrated framework for fast joint-space model-predictive control for reactive manipulation},
  author={Bhardwaj, Mohak and Sundaralingam, Balakumar and Mousavian, Arsalan and Ratliff, Nathan D and Fox, Dieter and Ramos, Fabio and Boots, Byron},
  booktitle={Conference on Robot Learning},
  pages={750--759},
  year={2022},
  organization={PMLR}
}

@inproceedings{inproceedings_bangura,
author = {Bangura, Moses},
year = {2014},
month = {08},
pages = {11773-11780},
title = {Real-Time Model Predictive Control for Quadrotors},
volume = {47},
journal = {IFAC Proceedings Volumes},
doi = {10.3182/20140824-6-ZA-1003.00203}
}

@INPROCEEDINGS{7029990_erez,
  author={Erez, Tom and Lowrey, Kendall and Tassa, Yuval and Kumar, Vikash and Kolev, Svetoslav and Todorov, Emanuel},
  booktitle={2013 13th IEEE-RAS International Conference on Humanoid Robots (Humanoids)}, 
  title={An integrated system for real-time model predictive control of humanoid robots}, 
  year={2013},
  volume={},
  number={},
  pages={292-299},
  doi={10.1109/HUMANOIDS.2013.7029990}}

@INPROCEEDINGS{5152817_ratliff,
  author={Ratliff, Nathan and Zucker, Matt and Bagnell, J. Andrew and Srinivasa, Siddhartha},
  booktitle={2009 IEEE International Conference on Robotics and Automation}, 
  title={CHOMP: Gradient optimization techniques for efficient motion planning}, 
  year={2009},
  volume={},
  number={},
  pages={489-494},
  doi={10.1109/ROBOT.2009.5152817}}

@inproceedings{inproceedings_schulman_ho_jonathan_lee,
author = {Schulman, John and Ho, Jonathan and Lee, Alex and Awwal, Ibrahim and Bradlow, Henry and Abbeel, Pieter},
year = {2013},
month = {06},
pages = {},
title = {Finding Locally Optimal, Collision-Free Trajectories with Sequential Convex Optimization},
doi = {10.15607/RSS.2013.IX.031}
}

@INPROCEEDINGS{8793889_qureshi,
  author={Qureshi, Ahmed H. and Simeonov, Anthony and Bency, Mayur J. and Yip, Michael C.},
  booktitle={2019 International Conference on Robotics and Automation (ICRA)}, 
  title={Motion Planning Networks}, 
  year={2019},
  volume={},
  number={},
  pages={2118-2124},
  doi={10.1109/ICRA.2019.8793889}}

@inproceedings{fishman2023motion,
  title={Motion policy networks},
  author={Fishman, Adam and Murali, Adithyavairavan and Eppner, Clemens and Peele, Bryan and Boots, Byron and Fox, Dieter},
  booktitle={Conference on Robot Learning},
  pages={967--977},
  year={2023},
  organization={PMLR}
}

@inproceedings{janner2022planning,
  title = {Planning with Diffusion for Flexible Behavior Synthesis},
  author = {Michael Janner and Yilun Du and Joshua Tenenbaum and Sergey Levine},
  booktitle = {International Conference on Machine Learning},
  year = {2022},
}

@INPROCEEDINGS{10610013_dipper,
  author={Liu, Jianwei and Stamatopoulou, Maria and Kanoulas, Dimitrios},
  booktitle={2024 IEEE International Conference on Robotics and Automation (ICRA)}, 
  title={DiPPeR: Diffusion-based 2D Path Planner applied on Legged Robots}, 
  year={2024},
  volume={},
  number={},
  pages={9264-9270},
  doi={10.1109/ICRA57147.2024.10610013}}

@INPROCEEDINGS{carvalho2023motion,
  author={Carvalho, João and Le, An T. and Baierl, Mark and Koert, Dorothea and Peters, Jan},
  booktitle={2023 IEEE/RSJ International Conference on Intelligent Robots and Systems (IROS)}, 
  title={Motion Planning Diffusion: Learning and Planning of Robot Motions with Diffusion Models}, 
  year={2023},
  volume={},
  number={},
  pages={1916-1923},
  doi={10.1109/IROS55552.2023.10342382}}

@inproceedings{sohl2015deep,
  title={Deep unsupervised learning using nonequilibrium thermodynamics},
  author={Sohl-Dickstein, Jascha and Weiss, Eric and Maheswaranathan, Niru and Ganguli, Surya},
  booktitle={International conference on machine learning},
  pages={2256--2265},
  year={2015},
  organization={PMLR}
}

@article{ho2020denoising,
  title={Denoising diffusion probabilistic models},
  author={Ho, Jonathan and Jain, Ajay and Abbeel, Pieter},
  journal={Advances in neural information processing systems},
  volume={33},
  pages={6840--6851},
  year={2020}
}

@inproceedings{strub2020adaptively,
  title={Adaptively Informed Trees (AIT): Fast Asymptotically Optimal Path Planning through Adaptive Heuristics},
  author={Strub, Marlin P and Gammell, Jonathan D},
  booktitle={2020 IEEE International Conference on Robotics and Automation (ICRA)},
  pages={3191--3198},
  year={2020},
  organization={IEEE}
}

@article{xie2021geometric,
  title={Geometric Fabrics for the Acceleration-based Design of Robotic Motion},
  author={Mandy Xie and Karl Van Wyk and Anqi Li and Muhammad Asif Rana and Dieter Fox and Byron Boots and Nathan D. Ratliff},
  journal={ArXiv},
  year={2020},
  volume={abs/2010.14750},
  url={https://api.semanticscholar.org/CorpusID:225094405}
}

@MISC{coumans2021,
author =   {Erwin Coumans and Yunfei Bai},
title =    {PyBullet, a Python module for physics simulation for games, robotics and machine learning},
howpublished = {\url{http://pybullet.org}},
year = {2016--2021}
}

@misc{tamp_yang2023planning,
      title={Planning as In-Painting: A Diffusion-Based Embodied Task Planning Framework for Environments under Uncertainty}, 
      author={Cheng-Fu Yang and Haoyang Xu and Te-Lin Wu and Xiaofeng Gao and Kai-Wei Chang and Feng Gao},
      year={2023},
      eprint={2312.01097},
      archivePrefix={arXiv},
      primaryClass={cs.CV}
}

@inproceedings{tamp_yang2023diffusion,
    title={{Compositional Diffusion-Based Continuous Constraint Solvers}},
    author={Yang, Zhutian and Mao, Jiayuan and Du, Yilun and Wu, Jiajun and Tenenbaum, Joshua B. and Lozano-P{\'e}rez, Tom{\'a}s and Kaelbling, Leslie Pack},
    booktitle={Conference on Robot Learning},
    year={2023},
  }

@inproceedings{tamp_structdiffusion2023,
    title     = {StructDiffusion: Language-Guided Creation of Physically-Valid Structures using Unseen Objects},
    author    = {Liu, Weiyu and Du, Yilun and Hermans, Tucker and Chernova, Sonia and Paxton, Chris},
    year      = {2023},
    booktitle = {RSS 2023}
}

@inproceedings{tamp_fang2023dimsam,
    title={Di{MS}am: Diffusion Models as Samplers for Task and Motion Planning under Partial Observability},
    author={Xiaolin Fang and Caelan Garrett and Clemens Eppner and Tom{\'a}s Lozano-P{\'e}rez and Leslie Kaelbling and Dieter Fox},
    booktitle={CoRL 2023 Workshop on Learning Effective Abstractions for Planning (LEAP)},
    year={2023},
    url={https://openreview.net/forum?id=a14qioqpel}
}

@article{tamp_chang2023denoising,
  title={Denoising Heat-inspired Diffusion with Insulators for Collision Free Motion Planning},
  author={Chang, Junwoo and Ryu, Hyunwoo and Kim, Jiwoo and Yoo, Soochul and Seo, Joohwan and Prakash, Nikhil and Choi, Jongeun and Horowitz, Roberto},
  journal={arXiv preprint arXiv:2310.12609},
  year={2023}
}

@inproceedings{pose_suresh2022midastouch,
    title={{M}idas{T}ouch: {M}onte-{C}arlo inference over distributions across sliding touch},
    author={Suresh, Sudharshan and Si, Zilin and Anderson, Stuart and Kaess, Michael and Mukadam, Mustafa},
    booktitle = {Proc. Conf. on Robot Learning, CoRL},
    address = {Auckland, NZ},
    month = dec,
    year = {2022}
}

@article{pose_simeonov2023rpdiff,
                author = {Simeonov, Anthony
                            and Goyal, Ankit
                            and Manuelli, Lucas
                            and Yen-Chen, Lin
                            and Sarmiento, Alina,
                            and Rodriguez, Alberto
                            and Agrawal, Pulkit
                            and Fox, Dieter},
                title = {Shelving, Stacking, Hanging: Relational
                            Pose Diffusion for Multi-modal Rearrangement},
                journal={Conference on Robot Learning},
                year={2023}
            }

@inproceedings{grasping_urain2023se,
  title={Se (3)-diffusionfields: Learning smooth cost functions for joint grasp and motion optimization through diffusion},
  author={Urain, Julen and Funk, Niklas and Peters, Jan and Chalvatzaki, Georgia},
  booktitle={2023 IEEE International Conference on Robotics and Automation (ICRA)},
  pages={5923--5930},
  year={2023},
  organization={IEEE}
}

@InProceedings{grasping_xu2023unidexgrasp,
    author    = {Xu, Yinzhen and Wan, Weikang and Zhang, Jialiang and Liu, Haoran and Shan, Zikang and Shen, Hao and Wang, Ruicheng and Geng, Haoran and Weng, Yijia and Chen, Jiayi and Liu, Tengyu and Yi, Li and Wang, He},
    title     = {UniDexGrasp: Universal Robotic Dexterous Grasping via Learning Diverse Proposal Generation and Goal-Conditioned Policy},
    booktitle = {Proceedings of the IEEE/CVF Conference on Computer Vision and Pattern Recognition (CVPR)},
    month     = {June},
    year      = {2023},
    pages     = {4737-4746}
}

@INPROCEEDINGS{robot_learning_li2023crossway,
  author={Li, Xiang and Belagali, Varun and Shang, Jinghuan and Ryoo, Michael S.},
  booktitle={2024 IEEE International Conference on Robotics and Automation (ICRA)}, 
  title={Crossway Diffusion: Improving Diffusion-based Visuomotor Policy via Self-supervised Learning}, 
  year={2024},
  volume={},
  number={},
  pages={16841-16849},
  doi={10.1109/ICRA57147.2024.10610175}}

@INPROCEEDINGS{9636651_lydia,
  author={Moll, Mark and Chamzas, Constantinos and Kingston, Zachary and Kavraki, Lydia E.},
  booktitle={2021 IEEE/RSJ International Conference on Intelligent Robots and Systems (IROS)}, 
  title={HyperPlan: A Framework for Motion Planning Algorithm Selection and Parameter Optimization}, 
  year={2021},
  volume={},
  number={},
  pages={2511-2518},
  doi={10.1109/IROS51168.2021.9636651}}

@INPROCEEDINGS{844730,
  author={Kuffner, J.J. and LaValle, S.M.},
  booktitle={Proceedings 2000 ICRA. Millennium Conference. IEEE International Conference on Robotics and Automation. Symposia Proceedings (Cat. No.00CH37065)}, 
  title={RRT-connect: An efficient approach to single-query path planning}, 
  year={2000},
  volume={2},
  number={},
  pages={995-1001 vol.2},
  doi={10.1109/ROBOT.2000.844730}}


\end{document}